\documentclass[acmsmall,screen]{acmart}
\AtBeginDocument{%
  }

\setcopyright{acmlicensed}
\usepackage{booktabs}
\usepackage{multirow}
\usepackage[most]{tcolorbox}
\usepackage{xcolor}
\usepackage{float}
\newtcolorbox{promptbox}[1]{
  colback=white,
  colframe=gray!60,
  coltitle=white,
  colbacktitle=gray!60,
  title={#1},
  fonttitle=\bfseries\small,
  boxrule=0.6pt,
  arc=3pt,
  left=8pt,
  right=8pt,
  top=6pt,
  bottom=6pt,
  boxed title style={
    top=2pt,
    bottom=2pt,
    left=8pt,
    right=8pt,
    boxrule=0pt,
    arc=3pt
  }
}
\usepackage[ruled,vlined]{algorithm2e}
\definecolor{eventblue}{RGB}{0,114,178}
\definecolor{evidencegreen}{RGB}{0,150,120}
\definecolor{docred}{RGB}{213,94,0}
\definecolor{rulepurple}{RGB}{117,112,179}

\newcommand{\EventTag}{\textcolor{eventblue}{\texttt{<Event>}}}
\newcommand{\EvidenceTag}{\textcolor{evidencegreen}{\texttt{<Current Evidence>}}}
\newcommand{\DocsTag}{\textcolor{docred}{\texttt{<Documents>}}}
\newcommand{\RuleTag}{\textcolor{rulepurple}{\texttt{<Rulebook>}}}

\definecolor{positivegreen}{RGB}{0,140,80}
\definecolor{negativered}{RGB}{180,40,40}

\newcommand{\PositiveTag}{\textcolor{positivegreen}{\texttt{<Positive Cases>}}}
\newcommand{\NegativeTag}{\textcolor{negativered}{\texttt{<Negative Cases>}}}
\definecolor{articlebrown}{RGB}{150,90,30}
\newcommand{\ArticleTag}{\textcolor{articlebrown}{\texttt{<Article>}}}
\newcommand{\TimelineTag}{\textcolor{teal}{\texttt{<Timeline>}}}
\newcommand{\FeedbackTag}{\textcolor{purple}{\texttt{<Feedback>}}}
\begin{document}
  \title[Agent-Driven Evidence Selection for Temporally Disordered Fake News Detection]{Beyond the Evidence Coherence Trap: Agent-Driven Evidence Selection for Temporally
  Disordered Fake News Detection}

\author{Chi Wang}
\affiliation{%
  \institution{Chongqing University}
  \city{Chongqing}
  \country{China}}
\email{wangchi@stu.cqu.edu.cn}
\orcid{0009-0001-7835-498X}

\author{Min Gao}
\authornote{Corresponding author.}
\email{webmaster@marysville-ohio.com}
\affiliation{%
  \institution{Chongqing University}
  \city{Chongqing}
  \country{China}}
\email{gaomin@cqu.edu.cn}
\orcid{0000-0003-0127-7477}
\additionalaffiliation{%
  \institution{Key Laboratory of Dependable Service Computing in Cyber Physical Society (Chongqing University), Ministry of Education of China}%
}

\author{Yifei Zhang}
\affiliation{%
  \institution{Chongqing University}
  \city{Chongqing}
  \country{China}} 
\email{202624100951@stu.cqu.edu.cn}
\orcid{0009-0008-0084-1068}

\author{Bincheng Gu}
\affiliation{%
  \institution{Chongqing University}
  \city{Chongqing}
  \country{China}}
\email{gubincheng@stu.cqu.edu.cn}
\orcid{0009-0006-1734-299X}

\author{Zongwei Wang}
\affiliation{%
  \institution{Chongqing University}
  \city{Chongqing}
  \country{China}} 
\email{zongwei@cqu.edu.cn}
\orcid{0009-0003-2606-9508}

\author{Wentao Li}
\affiliation{%
  \institution{University of Leicester}
  \city{Leicester}
  \country{England}}
\orcid{0000-0003-4941-8814}


\begin{abstract}

Fake news undermines public trust and the integrity of online information ecosystems. Temporally disordered fake news is especially elusive because it rearranges factual events into a false chronology. Individual events may remain factually valid even when their temporal relations are misrepresented. When similar events involving the same entity recur, retrieval may return semantically plausible evidence from the wrong period. Consequently, independent event verification may rely on locally coherent but temporally mismatched evidence and incorrectly validate the article. We call this the \textit{evidence coherence trap}. Because existing benchmarks rarely isolate this setting, we construct ChronoFake, a source-disjoint benchmark of 30,706 articles derived from temporally anchored real events. It distinguishes temporal manipulation from valid narration, including flashbacks. To address this, we propose DelSA, a \underline{D}ocument S\underline{el}ection \underline{S}elf-Updating \underline{A}gent that iteratively refines evidence selection during training and applies a frozen rulebook for one-pass evidence selection at inference. DelSA uses an LLM to characterize candidate evidence, while a verifier-trained reranker balances local relevance against timeline consistency. Finally, the classifier compares the article-implied chronology extracted by the LLM with the evidence-derived chronology to reveal global temporal conflicts. DelSA outperforms existing methods on ChronoFake, while remaining competitive on PolitiFact and GossipCop. ChronoFake is available at \url{https://anonymous.4open.science/r/ChronoFake-07C2.}

\end{abstract}




\maketitle

\section{Introduction}

Fake news has become a persistent threat to the integrity of online information ecosystems~\cite{[1]}. Beyond fabricated events, misinformation can also be constructed from real information organized into misleading narratives~\cite{[47],[48]}. Temporal fact-verification studies further show that a claim may become false when the temporal relations among otherwise factual events are misrepresented~\cite{[2],[32]}. For example, one narrative claims that Adam Schiff gave his account of President Trump's call with the Ukrainian president before the White House released the call memorandum.\footnote{\url{https://www.factcheck.org/2019/11/trump-muddies-impeachment-timeline/}} Both events occurred, but the memorandum was released one day before Schiff's account. The falsehood therefore lies in the asserted temporal relation rather than in either event itself. Detecting such manipulation is difficult because checking each event in isolation may confirm all of them while leaving the misleading chronology unchallenged~\cite{[2],[31],[32]}. Large language models (LLMs) further increase this risk by making misleading narratives easier to generate at scale~\cite{[9],[28]}. Effective detection must therefore verify how factual events are temporally organized across the article instead of relying only on event-level factuality.

\begin{figure}[htbp]      
  \centering              
  \includegraphics[width=\linewidth]{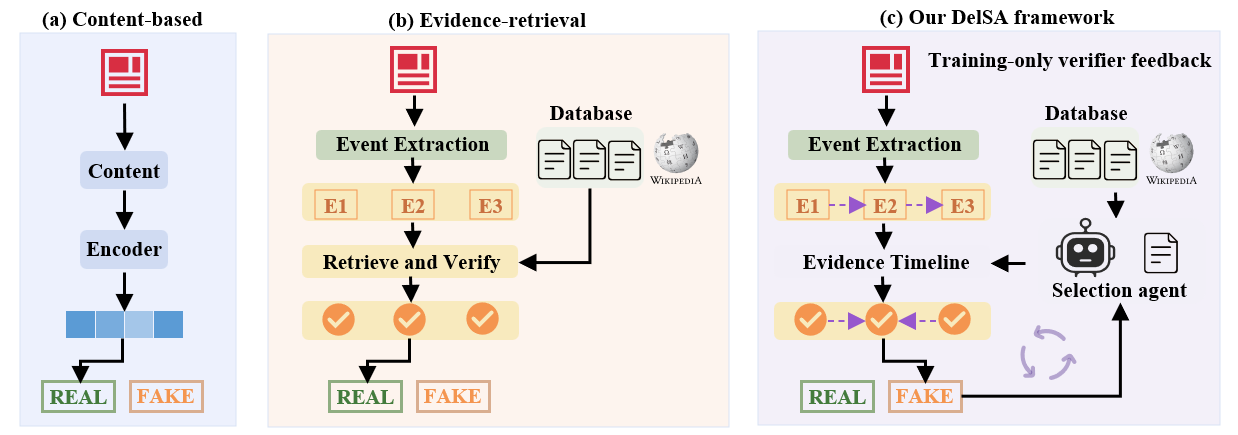}
  \caption{Comparison of detection paradigms for fake news. (a) Content-based methods rely on text semantics and ignore temporal structure.
(b) Evidence-retrieval methods verify each event independently and may predict the article as real when all events are locally supported, but miss global
chronological inconsistency. (c) Our DelSA framework leverages a self-updating agent to guide evidence reselection, reconstruct
the evidence-grounded timeline, and detect temporal conflicts.}   
  \label{fig:1}        
\end{figure}

Many fake news detectors estimate veracity from content-derived representations~\cite{[5],[6],[25]}. Recent LLM-enhanced methods improve robustness to LLM-generated or stylistically altered news~\cite{[9],[29]}. Nevertheless, temporally disordered fake news can preserve factual plausibility and semantic coherence because its component events are real. As shown in Fig.~\ref{fig:1}(a), content-based detectors may therefore overlook a falsehood expressed through incorrect temporal relations among events. Retrieval-augmented methods address this limitation by grounding verification in external evidence~\cite{[12],[51],[52]}. However, they often assess each event against its evidence in isolation. In the impeachment example, valid evidence can be retrieved separately for Schiff's account and the release of the call memorandum. A local verifier may thus confirm both events without comparing the chronology supported by the evidence with that asserted in the article, as illustrated in Fig.~\ref{fig:1}(b). We call this failure the \textit{evidence coherence trap}: locally consistent evidence creates the appearance of factual support while leaving the article-level chronology unverified.

Escaping the evidence coherence trap requires evidence selection to consider the temporal context of the article as a whole. When similar events involving the same entity occur at different times, semantic similarity alone cannot distinguish candidate documents associated with different stages of the same story. Recent methods incorporate temporal information into evidence retrieval~\cite{[32]} or reconstruct evidence-grounded timelines for claim verification~\cite{[2]}. However, neither approach explicitly uses article-level chronological assessment to revise previously selected evidence. The problem becomes harder when event mentions lack explicit dates. In such cases, the appropriate evidence for an event may become clear only after its position relative to other events is inferred, yet inferring that position also depends on the selected evidence. This interdependence motivates a mechanism that uses the current evidence-grounded timeline to revise evidence assignments and converts downstream verification feedback into reusable selection knowledge.

A rigorous evaluation of temporally disordered fake news detection should isolate temporal manipulation from other cues of falsity. Otherwise, a detector may exploit content artifacts rather than infer the underlying chronology. We therefore construct ChronoFake, a controlled benchmark from event units grounded in Wikipedia evidence. All event contents remain factual, and labels are determined by whether the temporal relations expressed in an article are consistent with the evidence-grounded chronology. The benchmark contains temporally manipulated false articles together with sequential and flashback true articles. Flashback articles present events nonlinearly while preserving valid temporal relations, preventing surface order alone from determining the label. ChronoFake therefore complements real-world benchmarks by providing a focused evaluation of temporal narrative verification.

Building on this task, we propose DelSA (\underline{D}ocument s\underline{el}ection \underline{S}elf-Updating \underline{A}gent), which couples time-aware evidence selection with article-level contextual timeline reasoning. Rather than equating textual mention order with chronology, DelSA uses an LLM to infer the temporal order asserted by the article from explicit time expressions, discourse cues, and relations among surrounding events. It therefore preserves the valid chronology conveyed by a legitimate flashback while exposing the incompatible chronology asserted by a temporally manipulated narrative. As shown in Fig.~\ref{fig:1}(c), DelSA iteratively reselects evidence for temporally ambiguous events during training, prioritizing documents that are not only semantically relevant to the event itself but also temporally consistent with the article-implied event context. DelSA further maintains a natural-language rulebook that is updated during training with downstream verification feedback: successful updates are summarized into reusable selection rules, while failed updates are converted into warnings or avoid-patterns. This allows DelSA to accumulate evidence-selection experience and better handle multi-time-point ambiguity in future samples.

The main contributions of this work are as follows:

\begin{itemize}

    \item To the best of our knowledge, we are the first to study the article-level detection of temporally disordered fake news, in which individually factual events are arranged into a misleading chronology. We further identify the evidence coherence trap, where locally valid evidence obscures article-level chronological inconsistency.

    \item We introduce ChronoFake, a controlled benchmark constructed from evidence-grounded event units. It contains temporally manipulated false articles together with sequential and flashback true articles, enabling a focused evaluation of temporal narrative verification.

    \item We propose DelSA, a rule-guided self-updating agent. During training, it refines evidence assignments using the current evidence-grounded timeline as context and converts downstream verification feedback into reusable selection rules. Experiments show that DelSA outperforms strong baselines on ChronoFake while remaining competitive on conventional fake news benchmarks.
\end{itemize}

\section{Related Work}
\subsection{Fake News Detection}

\subsubsection{Content-Based Fake News Detection.}
Early fake news detection studies mainly formulate the task as a supervised classification problem, where models learn discriminative patterns from news content, including lexical features, syntactic structures, writing styles, and document-level semantics~\cite{[5]}. With the development of neural representation learning, hierarchical and pretrained language models have been widely used to capture richer textual representations. For example, HAN models news articles with hierarchical attention over words and sentences, while BERT-based detectors fine-tune pretrained contextual representations for veracity prediction~\cite{[23],[24]}. Beyond plain text classification, dEFEND introduces an explainable detection framework by jointly modeling news content and user comments~\cite{[8]}. More recent methods further enhance content modeling with graph-based semantic interaction, dynamic representation selection as represented by BREAK and DRES~\cite{[6],[25]}.

In addition to textual content, multimodal fake news detection has become an important branch, since social media news often contains both text and images. CAFE learns cross-modal ambiguity between textual and visual modalities to detect inconsistent multimodal signals~\cite{[35]}, while HMCAN designs hierarchical multimodal contextual attention to learn text-image representations~\cite{[36]}. CSFND studies contextual semantic representation learning for multimodal fake news detection~\cite{[37]}, and ALGM introduces a generalized deep Markov random field framework to model multimodal ambiguity~\cite{[38]}. More recently, MIMoE-FND uses a mixture-of-experts architecture to model modality agreement and semantic alignment for robust multimodal detection~\cite{[39]}. However, these methods mainly focus on content-level semantic, stylistic, emotional, or multimodal inconsistencies, while our task targets chronological misinformation where individual events may be factual but their temporal organization is misleading.

\subsubsection{LLM-Based Fake News Detection.}
Large language models have recently been introduced into fake news detection due to their strong semantic understanding, reasoning, and explanation abilities. Existing studies use LLMs in different ways: ARG~\cite{[26]} explores how LLMs can act as advisors by providing rationales for downstream fake news detection; L-Defense~\cite{[46]} and GenFEND~\cite{[27]} leverage LLM-generated evidence, explanations, or knowledge to enhance veracity prediction; SheepDog~\cite{[28]} studies robust fake news detection under LLM-empowered style attacks; Adstyle~\cite{[29]} improves robustness by using LLM-based adversarial style augmentation and LIFE~\cite{[9]} enhances the ability to detect LLM-generated false information by introducing LLM inference signals. These methods demonstrate that LLMs can provide useful auxiliary signals for fake news detection, especially when surface-level textual patterns are insufficient.

Beyond using LLMs as direct predictors or rationale generators, some studies incorporate external knowledge into LLM-based verification. For example, FactLLaMA~\cite{[40]} optimizes instruction-following language models with external knowledge for automated fact-checking. This line of work suggests that LLM-based fake news detection benefits from grounding model reasoning in external information rather than relying only on parametric knowledge. However, existing LLM-based detectors usually perform sample-independent reasoning or generate auxiliary signals in a one-shot manner, while our method uses an LLM-driven agent to accumulate reusable document-selection rules from previous temporal evidence selection failures.

\subsubsection{Evidence-based Fake News Detection.}
Evidence-based fake news detection aims to improve reliability and interpretability by retrieving external evidence for news verification. Compared with content-based methods, evidence-grounded approaches reduce the dependence on shallow textual patterns and verify news claims against external information sources.
Early evidence-aware systems such as DeClarE~\cite{[41]} aggregate evidence articles together with source trustworthiness for credibility assessment. Subsequent work has improved evidence interaction at different granularities. GET~\cite{[10]} represents claims and evidence as graphs to capture long-distance semantic dependencies and reduce redundancy in retrieved evidence.

Recent fake news detection systems have pushed this paradigm further toward iterative and explainable retrieval. MUSER~\cite{[11]} performs multi-step evidence retrieval to model dependencies among multiple evidence pieces. UEEI~\cite{[18]} first identifies suspicious fragments in the news, then conducts entity-level and relation-level retrieval, and finally applies multi-view coherence reasoning between suspicious fragments and candidate evidence. IMRRF~\cite{[12]} addresses evidence insufficiency and redundancy by combining multi-source retrieval with LLM-based evidence conversion and redundancy filtering. In adjacent real-world fact-verification settings, AVeriTeC~\cite{[42]} highlights the importance of evidence retrieval from the open web, while MADAM-RAG~\cite{[13]} shows that multi-agent debate can be effective when retrieved evidence is ambiguous, noisy, or conflicting.
Nevertheless, most existing evidence-based methods mainly optimize semantic relevance, evidence sufficiency, or propagation/contextual signals, while our work focuses on temporally ambiguous evidence selection, where semantically coherent evidence may still support an incorrect global chronology.

\subsection{Temporal Fact Verification}

Temporal reasoning has become an important direction in fact verification, as the veracity of many claims depends not only on factual consistency but also on when the claim and its evidence hold. Early work studies temporal evidence ranking for fact-checking. Time-aware Evidence Ranking~\cite{[19]} shows that evidence timestamps can substantially affect evidence relevance and improves verification for time-sensitive claims by incorporating temporal constraints into ranking. Implicit Temporal Reasoning for Evidence-Based Fact-Checking~\cite{[31]} further grounds claims and evidence on shared timelines constructed from publication dates and temporal expressions, demonstrating that temporal signals can benefit downstream evidence-based verification even when temporal reasoning is modeled implicitly.

Recent studies make temporal verification more explicit by modeling temporal claims, temporal evidence, or event timelines. Mori et al.~\cite{[43]} focus on economic claims and translate natural-language claims into temporal Datalog queries to verify them against time-series evidence. TemporalFC~\cite{[44]} verifies temporally scoped assertions over knowledge graphs by assessing both factual correctness and temporal validity. Time Matters~\cite{[32]} proposes an end-to-end temporal claim verification framework that retrieves temporally relevant evidence and uses LLM-based reasoning for final verification. ChronoFact~\cite{[2]} further constructs timelines from claim and evidence events and models event-level, token-level, and time-level interactions for temporal fact verification. TSVer~\cite{[33]} studies fact verification against time-series evidence and emphasizes temporal and numerical reasoning with real-world claims and curated time-series data.
In retrieval-augmented reasoning, DyG-RAG~\cite{[34]} introduces dynamic event units and time-aware event-graph retrieval, showing the value of temporally anchored event representations for evidence retrieval and reasoning.

However, existing temporal fact verification methods mainly focus on claim-level verification or structured temporal evidence, while our work studies article-level chronological misinformation where multiple factual events may be assembled into a misleading temporal narrative, requiring iterative evidence repair and rule-guided document selection under temporal ambiguity.

\section{Methodology}
\begin{figure}[htbp]      
  \centering              
  \includegraphics[width=\linewidth]{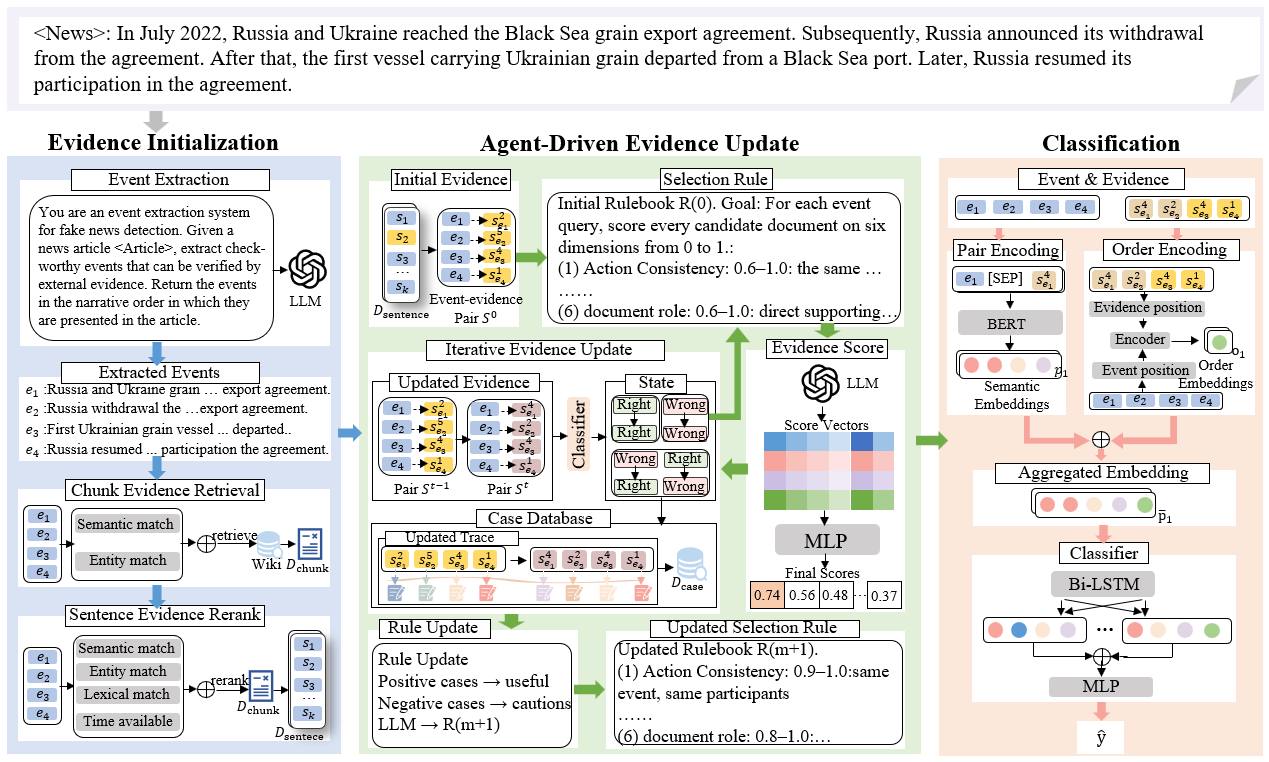}
  \caption{Overview of the DelSA framework for temporally disordered fake news detection. The input article is first decomposed into check-worthy events, and its asserted temporal relations are inferred by an LLM-based extractor. For each event, candidate evidence is retrieved from a Wikipedia-based corpus. During training, a time-aware, self-updating agent iteratively scores and refines evidence assignments using rule-guided evaluation, resolving temporal ambiguities and updating the rulebook based on downstream verification feedback. The classifier compares the article-implied chronology with the resulting evidence-grounded chronology for final veracity prediction.}   
  \label{fig:2}        
\end{figure}
This section begins with the task definition for temporally disordered fake news detection and an overview of the proposed Document Selection Self-Updating Agent (DelSA). It then presents the three main components of DelSA: the evidence initialization module (\S3.1), the agent-driven evidence update module (\S3.2), and the evidence-grounded classification module (\S3.3).

\textbf{\textit{Task Definition.}}
Given a news article $x$ with a gold veracity label $y$, we first decompose the article into a sequence of check-worthy events $E=(e_1,e_2,\ldots,e_n)$ and use the LLM to infer the temporal order asserted by the article. Let $\rho_x(i)$ denote the inferred temporal rank of event $e_i$ after resolving explicit time expressions and discourse cues, including flashback expressions. For each event $e_i$, a set of candidate evidence units $D_s^i=\{s_i^1,s_i^2,\ldots,s_i^K\}$ is retrieved from an external evidence corpus. Each candidate evidence unit $s_i^j$ may provide factual support, temporal clues, or misleadingly similar evidence for the event. The goal is to select temporally appropriate evidence for each event, reconstruct an evidence-grounded event chronology, compare it with the article-implied chronology, and predict an article-level label $\hat{y}$.

\textbf{\textit{Overview of DelSA.}}
DelSA addresses this task by coupling event-centric retrieval with agent-driven evidence update. As shown in Figure~\ref{fig:2}, DelSA first uses an LLM-based extractor to identify check-worthy events and infer the temporal relations asserted by the article, and retrieves candidate evidence for each event from a Wikipedia-based corpus. It then uses a self-updating agent to update temporally ambiguous evidence assignments. During training, DelSA further updates a rulebook using downstream verification feedback, enabling effective evidence-selection patterns to be reused in future samples. Finally, the classifier compares the article-implied chronology with the evidence-grounded chronology reconstructed from the updated event--evidence pairs.

\subsection{Evidence Initialization}
Following~\cite{[32]}, the purpose of this subsection is to extract the article-implied temporal structure and assign sentence-level evidence to each event. Given a news article $x$, DelSA first uses an LLM-based extractor to identify a sequence of check-worthy events $E=(e_1,e_2,\ldots,e_n)$ in textual mention order. In a separate output field, the extractor infers the temporal relations asserted by the article and produces the corresponding temporal ranks $\rho_x(i)$. These ranks are inferred from explicit dates and temporal discourse cues rather than from mention order alone; consequently, a legitimate flashback can have nonmonotonic mention positions while retaining a chronologically consistent inferred timeline. Each extracted event is a verifiable event mention that can be grounded in external evidence. The prompt template for event and temporal-relation extraction is shown below.

\begin{promptbox}{Event and Temporal-Relation Extraction Prompt Template}
\small
\setlength{\parskip}{0pt}
\setlength{\baselineskip}{10pt}
You are an event extraction system for fake news detection.

Given a news article \ArticleTag, extract check-worthy events that can be verified by external evidence.
Each extracted event must come directly from the article text.
Return the events in the narrative order in which they are presented in the article.
Infer the temporal relations asserted among the extracted events using explicit dates and discourse cues such as before, after, earlier, later, and retrospective expressions. Do not treat textual mention order alone as chronological order.
Skip opinions, vague background, rhetoric, and unverifiable claims.
Each event should have a clear subject.
Return only a valid JSON object containing the extracted event list, temporal relations, and inferred temporal ranks.
\end{promptbox}
For each event $e_i$, we first retrieve $K$ candidate chunks from a Wikipedia-based document corpus and then select $K$ evidence sentences after sentence-level reranking. This design ensures that the retrieved evidence is aligned in the form of a single event.

For chunk-level retrieval, to reduce boundary fragmentation, we first divide the Wikipedia corpus into fixed-length overlapping chunks. Each chunk is then encoded by BGE-M3~\cite{[45]} to obtain a dense semantic representation, and named entities are extracted from each chunk to form its entity set. Then, we build a FAISS index over the chunk embeddings and denote the resulting retrieval database as:
\begin{equation}
G=\{(c_j,\mathbf{g}_j,\mathcal{E}_j)\}_{j=1}^{M},
\end{equation}
where $c_j$ denotes the $j$-th chunk, $\mathbf{g}_j$ is its BGE-M3 embedding, $\mathcal{E}_j$ is the entity set extracted from the chunk, and $M$ is the total number of indexed chunks.

To improve the chance of recalling chunks that contain the gold supporting event, we employ semantic signals and entity signals. For semantic relevance, each event $e_i$ is encoded by BGE-M3 into an event embedding $\mathbf{h}_i$. The semantic similarity between $e_i$ and each indexed chunk $c_j$ is computed as:
\begin{equation}
V_1(e_i,c_j)=\cos(\mathbf{h}_i,\mathbf{g}_j),
\quad c_j\in G.
\end{equation}
Here, $V_1(e_i,c_j)$ measures whether the candidate chunk is semantically related to the event.

However, semantic retrieval alone may retrieve chunks that describe similar actions or event types but involve different entities. Therefore, we further introduce an entity alignment score to encourage the retrieved chunk to mention the same key entities as the target event. Let $\mathrm{Entities}(e_i)$ denote the entity set extracted from event $e_i$, and $\mathcal{E}_j$ denote the entity set associated with chunk $c_j$. The entity overlap score is defined as:
\begin{equation}
V_2(e_i,c_j)
=
\frac{
|\mathrm{Entities}(e_i)\cap \mathcal{E}_j|
}{
|\mathrm{Entities}(e_i)|
},
\end{equation}
A higher $V_2(e_i,c_j)$ indicates that the candidate chunk covers more event-specific entities, and therefore can further improve the quality of evidence recall.

Semantic relevance and entity alignment capture complementary retrieval signals. The former helps recall chunks that are close to the event in meaning, while the latter constrains the retrieval space to evidence involving the correct participants. Before weighted aggregation, all component scores are normalized to $[0,1]$ to ensure comparable scales. We combine these two signals to obtain the chunk-level retrieval score:
\begin{equation}
\mathrm{Score}_{\mathrm{c}}(e_i,c_j)
=
\alpha V_1(e_i,c_j)
+
\beta V_2(e_i,c_j),
\end{equation}
where $\alpha$ and $\beta$ control the relative contribution of semantic relevance and entity alignment. The top-$K$ chunks  under this score are used for event $e_i$:
\begin{equation}
D_c^{i}
=
\operatorname{TopK}^{K}_{c_j \in G}
\mathrm{Score}_{\mathrm{c}}(e_i,c_j).
\end{equation}

 Directly using the whole chunk as evidence may introduce noise into temporal verification. To obtain evidence that is more tightly aligned with the target event, we further perform sentence-level reranking over the retrieved chunks. Specifically, each chunk in $D_c^{i}$ is split into sentences, and all candidate sentences are collected as:
\begin{equation}
G_s^{i}
=
\{s_j ^k \mid s_j^k\in c_j\},
\end{equation}
where $s_j^k$ denotes the $k$-th sentence in chunk $c_j$.

For each candidate sentence $s_j^k$, we compute a sentence-level score by jointly considering semantic similarity, lexical overlap, entity consistency and temporal availability. The sentence-level score is defined as:
\begin{equation}
\begin{aligned}
\mathrm{Score}_{\mathrm{s}}(e_i,s_j ^k)
=&\;
\lambda_1 \cos\big(\mathbf{h}_i,\mathrm{BGE}(s_j ^k)\big)
+
\lambda_2 \mathrm{Ent}(e_i,s_j ^k) \\
&+
\lambda_3 \mathrm{Lex}(e_i,s_j ^k)
+
\lambda_4 \mathrm{Time}(e_i,s_j ^k),
\end{aligned}
\end{equation}
where $\mathrm{Ent}(e_i,s_j ^k)$ measures entity overlap between the event and the sentence, $\mathrm{Lex}(e_i,s_j ^k)$ measures lexical overlap, $\mathrm{Time}(e_i,s_j ^k)$ measures the temporal usefulness of the sentence. Similarly, all sentence-level component scores are normalized to $[0,1]$ before aggregation.

Finally, we select the top-$K$ ranked sentences as the sentence-level candidate evidence set:
\begin{equation}
D_s^i
=
\operatorname{TopK}_{s\in G_s^{i}}^{K}
\mathrm{Score}_{\mathrm{s}}(e_i,s).
\end{equation}
For each event $e_i$, the highest-ranked sentence is used as the initial evidence $s_i^{(0)}$, while the full candidate set $D_s^i$ is retained for subsequent evidence update. The initial evidence and timestamp are defined as:
\begin{equation}
s_i^{(0)}
=
\arg\max_{s\in D_s^i}
\mathrm{Score}_{\mathrm{s}}(e_i,s),
\quad
t_i^{(0)}=\tau(s_i^{(0)}),
\end{equation}
where $\tau(\cdot)$ is implemented with rule-based regular expressions, which extract explicit temporal expressions from the evidence sentence.

The resulting initialized event state is:
\begin{equation}
\mathcal{S}_i^{(0)}
=
\big(
e_i,\;
D_s^i,\;
s_i^{(0)},\;
t_i^{(0)}
\big).
\end{equation}
Thus, the article is transformed into a sequence of event-centric evidence states:
\begin{equation}
\mathcal{S}^{(0)}
=
\left(
\mathcal{S}_1^{(0)},
\mathcal{S}_2^{(0)},
\ldots,
\mathcal{S}_n^{(0)}
\right).
\end{equation}

\subsection{Agent-Driven Evidence Update}
\subsubsection{Rule-Guided Evidence Scoring}

After evidence initialization, each event $e_i$ is associated with a candidate evidence set
$D_s^i=\{s_i^1,s_i^2,\ldots,s_i^K\}$.
Although these candidates have been filtered by retrieval relevance, they may still be temporally or role-wise inappropriate for the target event, since different documents may describe similar events at different time points or with different participant roles.
Inspired by \cite{[53]}, DelSA incorporates a rule-guided evidence scoring mechanism for estimating the suitability of candidate evidence units for a given target event. The rule-guided mechanism encourages the scoring process to attend to semantic relevance, temporal consistency, and structural compatibility. The resulting evidence scores are then iteratively updated throughout the training process.

According to \cite{[54],[55]}, rather than using the LLM as an evidence selector, we employ it as a multi-dimensional semantic feature extractor. Specifically, at iteration $t$, given an event $e_i$, its current evidence $s_i^{(t)}$, the candidate evidence set $D_s^i$, the article-implied timeline $\rho_x$, and the current rulebook $R$, the LLM produces a structured quality vector for each candidate evidence:
\begin{equation}
\mathbf{v}_i^j
=
\mathrm{LLM}(e_i, s_i^{(t)}, D_s^i, \rho_x, R)
=
[
a_{\mathrm{act}}^j,
a_{\mathrm{role}}^j,
a_{\mathrm{sem}}^j,
a_{\mathrm{src}}^j,
a_{\mathrm{temp}}^j,
a_{\mathrm{doc}}^j
],
\end{equation}
where $a_{\mathrm{act}}^j$ checks whether the candidate describes the same core action as $e_i$;
$a_{\mathrm{role}}^j$ checks whether the participants keep the same subject-object roles;
$a_{\mathrm{sem}}^j$ measures overall semantic relevance;
$a_{\mathrm{src}}^j$ estimates whether the candidate sentence comes from a context that directly describes the target event, rather than from a loosely related background paragraph;
$a_{\mathrm{temp}}^j$ checks temporal compatibility with the target event;
and $a_{\mathrm{doc}}^j$ identifies whether the candidate serves as direct evidence, background context, related but insufficient information, or conflicting evidence. Each dimension is normalized into $[0,1]$.

The prompt template used for evidence scoring is shown below.

\begin{promptbox}{Rule-Guided Evidence Scoring Prompt Template}
\small
\setlength{\parskip}{0pt}
\setlength{\baselineskip}{10pt}
Given an event query \EventTag, its current evidence \EvidenceTag, a set of candidate evidence units \DocsTag, the article-implied timeline, and the current document selection rulebook \RuleTag, score each candidate evidence unit along the following dimensions: action consistency, role direction, semantic match, context quality, temporal alignment, and evidence role. Return only a JSON object containing the six normalized scores for every evidence unit. Do not directly choose or rank the evidence units; the downstream scorer will perform the final ranking.
\end{promptbox}

For each candidate evidence unit, DelSA uses a learnable MLP reranker to calculate the final score:
\begin{equation}
\ell^{j}_{i}=g_{\phi}(\mathbf{v}^{j}_{i}),\qquad \ell_i^j\in\mathbb{R},
\end{equation}

where $g_{\phi}$ is a lightweight multilayer perceptron that maps the six-dimensional quality vector into a scalar evidence-selection score. The candidate with the highest score is selected as the evidence for event \(e_i\):
\begin{equation}
  j^*=\arg\max_{1\leq j\leq K} \ell_i^j.
\end{equation}
Since hard evidence selection is non-differentiable, we train the reranker with verifier-derived weak supervision. Specifically, the downstream verifier converts the change in final classification loss before and after evidence update into a binary label, which is used to optimize the reranker.

Let \(z_i \in \{0,1\}\) denote the verifier-derived weak supervision label for the selected evidence of event \(e_i\). The reranker is optimized with a binary cross-entropy loss:
\begin{equation}
\mathcal{L}_{\mathrm{rerank}}
=
-\sum_i
\left[
z_i\log\sigma(\ell_i^{j^*})
+
(1-z_i)\log\left(1-\sigma(\ell_i^{j^*})\right)
\right],
\end{equation}
where \(\ell_i^{j^*}\) denotes the reranker score of the selected evidence candidate for event \(e_i\).

\subsubsection{Iterative Evidence Update}

Rule-guided evidence scoring provides a local operation for evaluating candidate evidence, but temporally disordered fake news requires article-level update.
The temporal suitability of an evidence unit is not determined only by its match to a single event; it also depends on whether the evidence is consistent with the evolving chronology formed by other event--evidence pairs.
Therefore, DelSA performs evidence update through an iterative loop.
At each update round, all events are reconsidered under the current evidence-grounded timeline, and their evidence assignments are updated together.
At update round $r$, DelSA maintains a sequence of event states $\mathcal{S}^{(r)}$.

The current evidence assignments induce an evidence-grounded timeline:
\begin{equation}
\mathcal{T}^{(r)}
=
\operatorname{Sort}
\left(
\left\{
(i,e_i,t_i^{(r)})
\right\}_{i=1}^{n}
\right),
\end{equation}
where events with valid timestamps are ordered chronologically.
The event identity $i$ is retained so that DelSA can align this evidence-grounded order with the LLM-inferred article chronology $\rho_x$.

During training, DelSA uses classifier feedback to control the update process.
After each update round, the prediction transition before and after evidence update is categorized into four states: right-to-right, wrong-to-wrong, wrong-to-right, and right-to-wrong.

Wrong-to-right indicates an effective evidence update.
The corresponding evidence-selection trajectory is stored as a positive memory for rulebook updating.
Right-to-wrong indicates a harmful evidence update, and the corresponding trajectory is stored as a negative memory.
Right-to-right is treated as a stable case and no further update is performed.
Wrong-to-wrong remains unresolved; DelSA continues the update with an additional feedback summary from the previous failed update.

For wrong-to-wrong cases, the next-round scorer receives a compact update history $H^{(r)}$, which records the previous evidence replacement, the resulting timestamp pattern, and the classifier outcome.
Given the current evidence-grounded timeline $\mathcal{T}^{(r)}$, DelSA performs timeline-aware evidence scoring as:
\begin{equation}
\ell_i^{j,(r)}
=
g_{\phi}
\left(
\mathrm{LLM}
\left(
e_i,\;
s_i^{(r)},\;
D_s^i,\;
\rho_x,\;
R,\;
\mathcal{T}^{(r)},\;
H^{(r)}
\right)
\right).
\end{equation}
Here, $e_i$ denotes the target event, $s_i^{(r)}$ denotes the currently selected evidence, $D_s^i$ denotes the candidate evidence set, $\rho_x$ denotes the article-implied timeline, $R$ denotes the current rulebook, $\mathcal{T}^{(r)}$ denotes the evidence-grounded timeline, and $H^{(r)}$ denotes the previous update feedback, which is derived from the transition between $\mathcal{S}^{(r-1)}$ and $\mathcal{S}^{(r)}$, including evidence replacements, timestamp changes, and the verifier outcome.

\begin{promptbox}{Timeline-Aware Evidence Scoring Prompt Template}
\small
\setlength{\parskip}{0pt}
\setlength{\baselineskip}{10pt}
Given an event query \EventTag, its current evidence \EvidenceTag, a set of candidate evidence units \DocsTag, the article-implied timeline, the current document selection rulebook \RuleTag, the evidence-grounded timeline \TimelineTag, and the previous update feedback \FeedbackTag, assign six normalized scores to each candidate evidence unit: action consistency, role direction, semantic match, context quality, temporal alignment, and evidence role. The two timelines provide temporal-stage context for distinguishing semantically similar evidence from different time points. The feedback summarizes the previous evidence update and verifier outcome. Return only a valid JSON object with the six scores for every evidence unit. Do not directly choose or rank the evidence units.
\end{promptbox}

The timeline $\mathcal{T}^{(r)}$ provides temporal-stage context for distinguishing temporally appropriate evidence from semantically similar candidates at different time points.
The feedback summary $H^{(r)}$ provides update-level context for avoiding previously failed evidence replacements.
All events are rescored under the same $\mathcal{T}^{(r)}$ and $H^{(r)}$, and the next evidence state is constructed synchronously after all events have been reconsidered.

The update terminates when the transition state is wrong-to-right, right-to-wrong, right-to-right, or when the maximum number of update rounds $\lambda$ is reached.
During validation and testing, the rulebook is frozen and DelSA performs one-pass rule-guided evidence selection using the article-implied timeline $\rho_x$ and the initialized evidence-grounded timeline, without label-dependent feedback or update history.

\subsubsection{Rulebook Updating}

The rule-guided scoring operation is controlled by a rulebook $R$, which defines how candidate evidence units are evaluated across different selection dimensions.
Since temporal evidence selection errors vary across entities, event types, and narrative structures, a fixed manually designed rulebook is insufficient.
DelSA therefore updates $R$ during training using the positive and negative memories collected from verifier-guided refinement.
The rulebook is updated only on the training split and is frozen during validation and testing.

At the end of each training epoch, DelSA aggregates the collected memories into a positive buffer $\mathcal{P}^{(m)}$ (wrong-to-right) and a negative buffer $\mathcal{N}^{(m)}$ (right-to-wrong) for epoch $m$.
Positive memories record evidence-selection trajectories that improve the verifier's prediction, while negative memories record harmful trajectories that degrade the prediction.
Each memory is represented as a compact update summary containing the event query, candidate evidence units, six-dimensional scores, selected evidence, timestamp pattern, and verifier outcome.

DelSA asks the LLM to abstract them into reusable evidence-selection principles.
Positive memories are used to reinforce effective selection patterns, while negative memories are used to introduce cautions against misleading evidence choices.
Raw articles and gold labels are not copied into the rulebook; only generalized selection rules are retained.

Formally, the rulebook is updated as:
\begin{equation}
R^{(m+1)}
=
U_{\mathrm{LLM}}
\left(
R^{(m)},\mathcal{P}^{(m)},\mathcal{N}^{(m)}
\right),
\end{equation}
where $R^{(m)}$ is the current rulebook, $\mathcal{P}^{(m)}$ and $\mathcal{N}^{(m)}$ are the positive and negative memory buffers collected in epoch $m$, and $U_{\mathrm{LLM}}$ denotes the LLM-based rule updating operation.

The prompt template for rulebook updating is shown below.

\begin{promptbox}{Feedback-Driven Rulebook Updating Prompt Template}
\small
\setlength{\parskip}{0pt}
\setlength{\baselineskip}{10pt}
You are maintaining a concise rulebook for evidence-unit reranking.

Given the current rulebook \RuleTag, positive update cases \PositiveTag, and negative update cases \NegativeTag, update the evidence selection rules.
The rulebook guides an LLM to assign six normalized scores in $[0,1]$ to each candidate evidence unit: action consistency, role direction, semantic match, context quality, temporal alignment, and document role.

Each positive case contains an event query, a short list of candidate evidence units, the six scores previously assigned by the LLM, the evidence unit selected by the scorer, and an outcome showing that this selection improved the final classifier.
Use positive cases to refine how each scoring dimension should be interpreted and calibrated.

Each negative case contains an update in which reranking changed an originally correct classification into an incorrect one.
Use negative cases only to add compact cautions, penalties, or boundary cases for the relevant scoring dimensions.

Keep the rulebook concise and organized under the six scoring dimensions.
Return only the updated rulebook.
\end{promptbox}

\subsection{Evidence-Grounded Classification}

After obtaining the updated event–evidence sequence, the final detection module predicts whether the article is fake by grounding each event in its selected evidence, rather than relying solely on surface text.
Specifically, given the updated state sequence:
\begin{equation}
\mathcal{S}^{(\ast)}
=
\left(
\mathcal{S}_1^{(\ast)},
\mathcal{S}_2^{(\ast)},
\ldots,
\mathcal{S}_n^{(\ast)}
\right),
\end{equation}

DelSA first constructs an evidence-grounded chronological order according to the timestamps extracted from the selected evidence:
\begin{equation}
\pi^{(\ast)}
=
\operatorname{argsort}_{i}
\left(
\left\{
t_i^{(\ast)}
\right\}_{i=1}^{n}
\right).
\end{equation}

For each event--evidence pair, DelSA constructs a pair-level representation using a pretrained language encoder:
\begin{equation}
\mathbf{p}_i
=
\mathrm{BERT}
\left(
[\mathrm{CLS}]\; e_i\; [\mathrm{SEP}]\; s_i^{(\ast)}\; [\mathrm{SEP}]
\right),
\end{equation}
where $\mathbf{p}_i$ captures whether the selected evidence semantically supports, contradicts, or only partially matches the corresponding event.

To make the detector aware of temporal narrative structure, DelSA further incorporates order information into each pair representation.
Let $\mathcal{T}^{(\ast)}(i)$ denote the chronological position of event $e_i$ in the evidence-grounded order, and let $\rho_x(i)$ denote its LLM-inferred temporal rank in the chronology asserted by the article.
We define an order embedding as:

\begin{equation}
\mathbf{o}_i
=
W_r\mathrm{PE}(
\mathcal{T}^{(\ast)}(i)-\rho_x(i)),
\end{equation}
where $\mathcal{T}^{(\ast)}(i)-\rho_x(i)$ measures the displacement between the article-implied temporal order and the evidence-grounded chronological order, $\mathrm{PE}(\cdot)$ is the sinusoidal positional encoding, and $W_r$ is a learnable projection matrix.
The order-aware event--evidence representation is then defined as:
\begin{equation}
\tilde{\mathbf{p}}_i
=
\left[
\mathbf{p}_i;
\mathbf{o}_i
\right].
\end{equation}
By comparing the LLM-inferred article chronology with the evidence-grounded chronology, DelSA provides the verifier with an explicit conflict signal while avoiding the assumption that a legitimate flashback must follow chronological mention order.

Following the evidence-grounded chronological order, the order-aware event--evidence representations are arranged into a sequence:
\begin{equation}
\mathbf{P}_{\pi}
=
\left[
\tilde{\mathbf{p}}_{\pi^{(\ast)}(1)},
\tilde{\mathbf{p}}_{\pi^{(\ast)}(2)},
\ldots,
\tilde{\mathbf{p}}_{\pi^{(\ast)}(n)}
\right].
\end{equation}
This sequence provides a temporally organized view of the article events after evidence update while retaining their article-implied temporal positions through the order embeddings.

To model the temporal dependency among event--evidence pairs, DelSA feeds $\mathbf{P}_{\pi}$ into a bidirectional LSTM:
\begin{equation}
\mathbf{Z}
=
\mathrm{BiLSTM}
\left(
\mathbf{P}_{\pi}
\right),
\end{equation}
where $\mathbf{Z}=[\mathbf{z}_1,\mathbf{z}_2,\ldots,\mathbf{z}_n]$ denotes the contextualized sequence representations.
The BiLSTM enables the model to capture how the evidence-grounded events evolve over time and whether their temporal organization is consistent with the article-level narrative.

The sequence representation is then aggregated into an article-level vector using attention pooling:
\begin{equation}
\alpha_i
=
\frac{
\exp\left(\mathbf{u}^{\top}\tanh(\mathbf{W}_a\mathbf{z}_i)\right)
}{
\sum_{k=1}^{n}
\exp\left(\mathbf{u}^{\top}\tanh(\mathbf{W}_a\mathbf{z}_k)\right)
},
\quad
\mathbf{z}
=
\sum_{i=1}^{n}
\alpha_i\mathbf{z}_i.
\end{equation}
The attention pooling layer allows the verifier to assign larger weights to event--evidence pairs that are more indicative of temporal inconsistency or misinformation.

Finally, the article-level representation is fed into a classification head:
\begin{equation}
\hat{\mathbf{y}} = \mathrm{MLP}(\mathbf{z})
\end{equation}
where $\hat{\mathbf{y}}$ denotes the predicted probability distribution over the real and fake labels.

The verifier is trained with the standard cross-entropy loss:
\begin{equation}
\mathcal{L}_{cls} = -\sum_{c\in \{0,1\}} y_c \log \hat{y}_c .
\end{equation}
The full training objective combines fake-news classification and reranker supervision:
\begin{equation}
\mathcal{L}
=
\mathcal{L}_{cls}
+
\eta \mathcal{L}_{\mathrm{rerank}},
\end{equation}
where \(\eta\) controls the contribution of reranker training.
The overall training procedure of DelSA is summarized in Algorithm~\ref{alg:delsa}.
\begin{algorithm}[t]
\caption{Overall Training Procedure of DelSA}
\label{alg:delsa}
\KwIn{
Training set $\mathcal{D}$, evidence corpus $\mathcal{G}$, initial rulebook $R^{(0)}$, maximum update rounds $\lambda$, loss weight $\eta$
}
\KwOut{
Trained classifier $F_{\theta}$, reranker $g_{\phi}$, and rulebook $R$
}

Initialize classifier $F_{\theta}$ and reranker $g_{\phi}$\;
Set $R \leftarrow R^{(0)}$\;

\For{each training epoch $m$}{
    Initialize positive memory $\mathcal{P}^{(m)}$ and negative memory $\mathcal{N}^{(m)}$\;

    \For{each article-label pair $(x,y)\in\mathcal{D}$}{
        Extract event sequence $E$ and article-implied temporal ranks $\rho_x$, and retrieve candidate evidence sets $\mathcal{D}_s$ from $\mathcal{G}$\;
        Initialize evidence state $\mathcal{S}^{(0)}$\;

        Perform iterative evidence update with $R$, $\lambda$, $F_{\theta}$, and $g_{\phi}$ to obtain final state $\mathcal{S}^{(*)}$\;
        Collect verifier feedback into $\mathcal{P}^{(m)}$, $\mathcal{N}^{(m)}$, or update history\;

        Construct evidence-grounded representations from $\mathcal{S}^{(*)}$\;
        Predict article label with verifier $F_{\theta}$\;

        Update $F_{\theta}$ and $g_{\phi}$ by minimizing
        $
        \mathcal{L}_{cls}+\eta\mathcal{L}_{rerank}
        $\;
    }

    Update rulebook:
    $
    R^{(m+1)}
    =
    U_{\mathrm{LLM}}
    \left(
    R^{(m)},\mathcal{P}^{(m)},\mathcal{N}^{(m)}
    \right)
    $\;
}

Freeze $R$ for validation and testing\;
Perform one-pass rule-guided evidence selection during inference\;

\Return $F_{\theta}$, $g_{\phi}$, and $R$\;
\end{algorithm}
\section{ChronoFake Benchmark Construction}

To investigate fake news caused by temporal disorder, we construct \textsc{ChronoFake}, a benchmark in which all event contents are grounded in real-world Wikipedia evidence, while falsehood is introduced by manipulating the temporal relations among events. Unlike conventional fake-news benchmarks that mainly focus on fabricated claims, semantic inconsistency, or stylistic deception, \textsc{ChronoFake} targets a subtler scenario: each event can be factually correct in isolation, but the article becomes misleading because the expressed temporal narrative conflicts with the true event chronology. Inspired by the event-centric dynamic-graph view of DyG-RAG~\cite{[34]}, we construct \textsc{ChronoFake} from temporally anchored event units and use entity-centered event graphs to support controlled chronology manipulation.

\subsection{Temporally Anchored Event Unit Construction}

Given a collection of Wikipedia-derived source documents $D=\{d_1,d_2,\ldots,d_n\}$ associated with a target entity, we first decompose the raw text into temporally grounded event units. Each Dynamic Event Unit (DEU) is defined as
\begin{equation}
\mathrm{DEU}=\{s_i,t_i,\mathrm{ID}_{event},\mathrm{ID}_{source}\},
\end{equation}
where $s_i$ denotes a self-contained factual event statement, $t_i$ is its normalized timestamp, and $\mathrm{ID}_{event}$ and $\mathrm{ID}_{source}$ denote the event identifier and source identifier, respectively. This event-level representation is necessary because paragraph-level chunks may contain multiple time points and blur event boundaries, making them unsuitable for controlled temporal manipulation.

Since ChronoFake is constructed from Wikipedia-derived event units, a naive article-level random split may introduce source leakage. To prevent this issue, we adopt a source-disjoint splitting protocol. We first group all DEUs by their $ID_{source}$, then split the source-document groups into training, validation, and test partitions. Article generation is performed within each partition, such that every generated article only uses DEUs whose $ID_{source}$ belongs to the corresponding partition. As a result, the same source document never contributes event units to multiple data splits.

For extraction, we segment each source document into overlapping chunks to preserve local context. Candidate event statements are then extracted from each chunk, and temporal expressions are parsed and normalized into explicit time anchors. Absolute dates are directly normalized, while relative expressions are resolved using the closest valid temporal reference in the local context. For interval expressions, we preserve the original span while using the earliest point for temporal indexing. We retain only informative eventive statements that contain named entities, state changes, or explicit outcomes, and apply sentence normalization and coreference resolution to ensure that each DEU is semantically coherent and temporally localized.

After DEU extraction, we organize event units associated with the same target entity into an entity-centered event graph $G=(V,E)$, where each node $v_i\in V$ corresponds to a DEU and each edge $e_{ij}\in E$ captures temporal-semantic relatedness between two event units. An edge is considered only when the two event units share at least one entity and occur within a temporal window $\Delta t$:
\begin{equation}
\mathrm{EntityOverlap}(v_i,v_j)>0
\quad \text{and} \quad
|t_i-t_j|\leq \Delta t.
\end{equation}
The edge weight is computed by combining entity overlap and temporal closeness:
\begin{equation}
\mathrm{sim}(v_i,v_j)=\frac{|\mathcal{E}_i\cap \mathcal{E}_j|}{|\mathcal{E}_i\cup \mathcal{E}_j|},
\end{equation}
\begin{equation}
w_{i,j}=\mathrm{sim}(v_i,v_j)\cdot \exp(-\alpha|t_i-t_j|),
\end{equation}
where $\mathcal{E}_i$ and $\mathcal{E}_j$ are the named-entity sets of $v_i$ and $v_j$, and $\alpha$ controls temporal decay. To reduce spurious links, we keep only the top-$K$ neighbors for each node:
\begin{equation}
N_i=\mathrm{Top}\text{-}K_{j\neq i}(w_{i,j}).
\end{equation}
The resulting graph defines local event neighborhoods in which chronology can be perturbed while preserving entity continuity and discourse plausibility.

\subsection{Temporal Narrative Generation}

For each target entity $e$, we collect its associated DEUs and construct the gold chronological trajectory by sorting timestamped event units in ascending temporal order. Let $\mathcal{V}_e=\{v_1,\ldots,v_m\}$ denote the DEUs linked to entity $e$. The gold trajectory is defined as
\begin{equation}
\sigma_e^*=\operatorname{Sort}\big(\{s_i\}_{v_i\in \mathcal{V}_e};\{t_i\}_{v_i\in \mathcal{V}_e}\big).
\end{equation}
This trajectory serves as the factual temporal backbone of the target entity.

Based on the gold trajectory, we generate three types of temporal narratives. First, \emph{sequential true articles} follow the gold chronological trajectory and present events in their actual temporal order. Second, \emph{flashback true articles} use a nonlinear surface narrative, such as starting from a later event and retrospectively referring to earlier events, while still preserving the correct temporal relations through explicit discourse cues. Third, \emph{temporally disordered fake articles} are generated by perturbing the temporal organization of graph-connected events while keeping their event contents unchanged.

Formally, let $\mathcal{P}(\cdot)$ denote a chronology perturbation operator. A manipulated trajectory is written as
\begin{equation}
\tilde{\sigma}_{e}=\mathcal{P}(\sigma_e^*),
\end{equation}
where $\mathcal{P}$ performs local reversal, cross-time insertion, or temporally inconsistent reordering over graph-connected event units. The perturbation is constrained by the event graph so that reordered events remain semantically coherent and entity-consistent. This avoids trivial negative samples caused by arbitrary shuffling and ensures that the resulting fake articles are misleading specifically because of temporal disorder rather than factual fabrication or lexical corruption.

\subsection{Label Definition and Quality Control}

Each final instance consists of a news article $d$ and a binary label $y\in\{0,1\}$. Instead of labeling an article according to whether its surface narrative order is identical to the gold chronological trajectory, we label it according to whether the temporal relations expressed by the article are consistent with the gold trajectory. This distinction is important because a legitimate flashback article may present events in a nonlinear order while still explicitly preserving the correct temporal relations.

Let $\mathcal{R}(d)$ denote the set of temporal relations expressed by article $d$, considering both surface event order and explicit discourse cues such as retrospective or flashback expressions. The label is defined as:
\begin{equation}
y=
\begin{cases}
0, & \text{if } \mathcal{R}(d) \text{ is consistent with } \sigma_e^*,\\
1, & \text{otherwise}.
\end{cases}
\end{equation}
Thus, $y=1$ corresponds to chronologically manipulated fake news, while $y=0$ includes both sequentially ordered true articles and flashback-style true articles whose expressed temporal relations remain faithful to the underlying chronology.

To ensure data quality, we retain only entities whose event trajectories contain multiple temporally distinguishable events and whose reordered narratives remain semantically interpretable after perturbation. We also filter out instances with missing or ambiguous timestamps, insufficient entity continuity, or incoherent generated narratives. Through this design, \textsc{ChronoFake} isolates temporal veracity from factual veracity: the event contents remain grounded in real evidence, while the misinformation arises from the article-level temporal organization.
\section{Experiments}

\subsection{Experimental Settings}
\subsubsection{Datasets}
\begin{table}[h]
\centering
\caption{Dataset statistics.}
\label{tab:datasets}
\begin{tabular}{lccc}
\toprule
Dataset & GossipCop & PolitiFact & ChronoFake \\
\midrule
Total News & 12,252 & 741 & 30,706  \\
Fake News  & 4,084  & 391 & 9,925  \\
Real News  & 8,168  & 350 & 20,781  \\
\bottomrule
\end{tabular}
\end{table}
We conduct experiments on three fake news detection benchmarks, including two widely used general fake news datasets and one temporally manipulated benchmark constructed for chronological misinformation detection.

\textbf{General Fake News Datasets.}
We use PolitiFact and GossipCop, two English fake news datasets derived from FakeNewsNet~\cite{[14]}. In our experiments, PolitiFact contains 741 news articles, and GossipCop contains 12,252 news articles. These two datasets are used to evaluate whether the proposed method can generalize beyond temporally manipulated samples and remain effective on conventional fake news detection benchmarks.

\textbf{Temporal Disorder Dataset.}
To specifically evaluate article-level temporal inconsistency detection, we use \textbf{ChronoFake}, a benchmark targeting temporally disordered disinformation. ChronoFake contains 30,706 instances, including 9,925 fake news articles, 10,390 sequentially ordered true news articles, and 10,391 flashback (nonlinear) true news articles (Table~\ref{tab:datasets}).

\subsubsection{Implementation Details}

We implement our framework in PyTorch and conduct all experiments on an NVIDIA GeForce RTX 4090 GPU. We set the maximum sequence length to 256, batch size to 8, and train the model for 10 epochs. The learning rates for BERT and task-specific parameters are set to \(2\times10^{-5}\) and \(1\times10^{-3}\), respectively. For temporal update, DelSA performs iterative evidence and timestamp updates for at most 5 rounds. The LLM backend is configured as gpt-5.4-mini. We report accuracy, weighted precision, weighted recall, and weighted F1 as evaluation metrics.
\subsubsection{Baselines}
To evaluate the effectiveness of DelSA, we compare it against a wide range of representative fake news detection models. To ensure a fair and rigorous comparison, all baseline methods are evaluated under the same experimental environment. Specifically, our comparison includes three categories of methods: (1) Content-Based models that rely primarily on news content for prediction; (2) LLM-Enhanced methods that leverage large language models to improve fake news detection; and (3) Retrieval-Augmented methods that explicitly retrieve and reason over external evidence.

The Content-Based fake news detection models are as follows:
\begin{itemize}
    \item \textbf{BERT}\cite{[24]}: a pre-trained language model fine-tuned for fake news detection.
    \item \textbf{HAN}\cite{[23]}: uses a hierarchical attention network to capture both word-level and sentence-level textual features for fake news detection.
    \item \textbf{dEFEND}\cite{[8]}: proposes an explainable fake news detection framework that exploits news content together with users' comments.
    \item \textbf{BREAK}\cite{[6]}: models news text and images with graph-based semantic interaction to improve fake news detection.
    \item \textbf{DRES}\cite{[25]}: identifies fake news based solely on text by dynamically selecting the most suitable textual representation and classifier ensemble for each news article according to its instance hardness.
\end{itemize}

In addition to the traditional content-based baselines, we further compare the following LLM-enhanced models:
\begin{itemize}
    \item \textbf{ARG}~\cite{[26]}: studies how large language models can assist fake news detection by providing rationales to guide downstream prediction.
    \item \textbf{L-Defense}~\cite{[46]}: leverages large language models to extract evidence and generate justifications for fake news detection.
    \item \textbf{GenFEND}~\cite{[27]}: enhances fake news detection with LLM-based knowledge generation and reasoning.
    \item \textbf{SheepDog}~\cite{[28]}: uses large language models to provide stronger semantic understanding and decision support for fake news detection.
    \item \textbf{LIFE}~\cite{[9]}: provides a paradigm for detecting fake news generated by LLM, utilizing LLM to reconstruct the features of news production.
\end{itemize}

Moreover, since our task focuses on evidence-grounded temporal verification, we additionally compare with the following retrieval-augmented fake news detection models:
\begin{itemize}
    \item \textbf{MUSER}\cite{[11]}: performs multi-step evidence retrieval and verification to improve fake news detection.
    \item \textbf{IMRRF}\cite{[12]}: enhances fake news detection through multi-source retrieval and evidence interaction modeling.
\end{itemize}

\subsection{Performance Comparison}
\begin{table*}[t]
\centering
\caption{Performance comparison on \textbf{PolitiFact} and \textbf{GossipCop}. Best results are in \textbf{bold}, and second best are \underline{underlined}.}
\label{tab:performance}
\begin{tabular}{l|l|cccc|cccc}
\toprule
\multirow{2}{*}{Category} & \multirow{2}{*}{Method}
& \multicolumn{4}{c|}{PolitiFact}
& \multicolumn{4}{c}{GossipCop} \\
\cline{3-10}
 &  & Acc. & Pre. & Rec. & F1
    & Acc. & Pre. & Rec. & F1 \\
\midrule
\multirow{5}{*}{Content-Based}
& HAN      & 0.724 & 0.743 & 0.724 & 0.692 & 0.734 & 0.722 & 0.734 & 0.709 \\
& dEFEND   & 0.821 & 0.830 & 0.821 & 0.823 & 0.849 & 0.846 & 0.849 & 0.847 \\
& BERT     & 0.850 & 0.850 & 0.850 & 0.849 & 0.828 & 0.820 & 0.828 & 0.801 \\
& BREAK    & 0.867 & 0.867 & 0.867 & 0.867 & 0.831 & 0.817 & 0.831 & 0.808 \\
& DRES     & 0.911 & 0.913 & 0.911 & 0.917 & \textbf{0.874} & \textbf{0.882} & \textbf{0.874} & \textbf{0.867} \\
\midrule
\multirow{5}{*}{LLM-Enhanced}
& ARG       & \underline{0.926} & \underline{0.924} & \underline{0.926} & \textbf{0.924} & 0.852 & 0.812 & 0.852 & 0.823 \\
& L-Defense & 0.899 & 0.902 & 0.899 & 0.896 & 0.862 & 0.858 & 0.862 & 0.861 \\
& GenFEND   & 0.899 & 0.897 & 0.900 & 0.898 & 0.850 & 0.853 & 0.850 & 0.848 \\
& SheepDog  & 0.898 & 0.900 & 0.898 & 0.899 & 0.843 & 0.831 & 0.843 & 0.834 \\
& LIFE    & 0.835 & 0.828 & 0.835 & 0.828 & 0.785 & 0.779 & 0.785 & 0.773 \\
\midrule
\multirow{2}{*}{Retrieval-Augmented}
& MUSER     & 0.746 & 0.740 & 0.746 & 0.741 & 0.758 & 0.758 & 0.758 & 0.758 \\
& IMRRF     & 0.724 & 0.783 & 0.724 & 0.732 & 0.741 & 0.762 & 0.741 & 0.747 \\
\midrule
\multirow{1}{*}{Our Method}
& DelSA & \textbf{0.932} & \textbf{0.936} & \textbf{0.932} & \textbf{0.924} & \underline{0.869} & \underline{0.877} & \underline{0.869} & \underline{0.863} \\
\bottomrule
\end{tabular}
\end{table*}
We benchmark the performance of \textbf{DelSA} against several baseline methods, including content-based, LLM-enhanced, and retrieval-augmented approaches. Experiments are conducted on two datasets: \textbf{PolitiFact} and \textbf{GossipCop}. The results are summarized in Table~\ref{tab:performance}, and key findings are as follows:

\begin{itemize}
    \item \textbf{DelSA achieves the best performance on PolitiFact and competitive performance on GossipCop.} As shown in Table~\ref{tab:performance}, DelSA achieves the highest accuracy, precision, recall, and ties for the best F1 on PolitiFact. On GossipCop, DelSA remains highly competitive with the strongest content-based baseline, showing only marginal differences across the main metrics.

    \item \textbf{DelSA outperforms traditional content-based baselines.} While these baselines rely primarily on surface-level textual or multimodal content features, DelSA leverages temporal reasoning and LLM-guided evidence selection to accurately resolve multi-time-point ambiguities. This highlights the importance of explicitly modeling event chronology and rule-guided document selection for fake news detection.

    \item \textbf{DelSA surpasses LLM-enhanced approaches.} Although LLM-enhanced baselines use large language models for auxiliary reasoning or semantic understanding, DelSA further exploits LLMs in a structured framework with self-updating rules and iterative evidence alignment. The results indicate that integrating LLM reasoning with temporal repair and rule-guided selection is more effective than relying on LLMs alone.

\end{itemize}
\subsection{Temporally Disordered Fake News Detection}
We benchmark the performance of \textbf{DelSA} against several baseline methods, including content-based, LLM-enhanced, and retrieval-augmented approaches. Experiments are conducted on ChronoFake. The results are summarized in Table~\ref{tab:disorder}, and key findings are as follows:

\begin{table*}[t]
\centering
\caption{Performance comparison on \textbf{ChronoFake}. Best results are in \textbf{bold}, and second best are \underline{underlined}.}
\label{tab:disorder}
\begin{tabular}{l|l|cccc}
\toprule
\multirow{2}{*}{Category} & \multirow{2}{*}{Method}
& \multicolumn{4}{c|}{ChronoFake} \\
\cline{3-6}
 &  & Acc. & Pre. & Rec. & F1  \\
\midrule
\multirow{5}{*}{Content-Based}
& HAN      & 0.672 & 0.452 & 0.672 & 0.540 \\
& dEFEND   & 0.681 & 0.463 & 0.681 & 0.551 \\
& BERT     & 0.797 & 0.763 & 0.797 & 0.765 \\
& BREAK    & 0.700 & 0.686 & 0.700 & 0.691 \\
& DRES     & \underline{0.819} & 0.817 & \underline{0.819} & \underline{0.818} \\

\midrule
\multirow{5}{*}{LLM-Enhanced}
& ARG       & 0.811 & 0.818 & 0.811 & 0.813 \\
& L-Defense & 0.687 & 0.644 & 0.687 & 0.600 \\
& GenFEND   & 0.684 & 0.746 & 0.684 & 0.694 \\
& SheepDog  & 0.816 & \underline{0.831} & 0.816 & 0.800 \\
& LIFE     & 0.663 & 0.777 & 0.662 & 0.715 \\
\midrule
\multirow{2}{*}{Retrieval-Augmented}
& MUSER     & 0.765 & 0.795 & 0.765 & 0.717 \\
& IMRRF     & 0.746 & 0.819 & 0.746 & 0.707 \\
\midrule
\multirow{2}{*}{Our Method}
& DelSA & \textbf{0.892} & \textbf{0.894} & \textbf{0.892} & \textbf{0.893} \\
& Imp.(\%)   & 8.913 & 7.581 & 8.913 & 9.169  \\
\bottomrule
\end{tabular}
\end{table*}

\begin{itemize}
    \item \textbf{DelSA achieves the best overall performance on ChronoFake.}
    As shown in Table~\ref{tab:disorder}, DelSA outperforms all baselines across accuracy, precision, recall, and F1. Compared with the best-performing baseline for each metric, DelSA improves the four metrics by 8.91\%, 7.58\%, 8.91\%, and 9.17\%, respectively, demonstrating its effectiveness in temporally inconsistent fake news detection.

    \item \textbf{Content-based methods are insufficient for temporal inconsistency modeling.}
    Although methods such as BERT, BREAK, and DRES can capture semantic and contextual features from news content, they mainly treat fake news detection as a static text classification task. Therefore, they may fail when fake news is caused not by obvious textual cues, but by conflicts among event timelines, claims, and supporting evidence.

    \item \textbf{LLM-enhanced and retrieval-augmented methods still lack temporal regulation.}
    LLM-enhanced methods benefit from stronger semantic reasoning, while retrieval-augmented methods introduce external evidence. However, they do not explicitly control temporal order or evidence consistency. As a result, LLM reasoning may be insufficiently grounded, and retrieved evidence may be noisy or temporally mismatched, limiting their performance on ChronoFake.

    \item \textbf{DelSA benefits from temporal repair and rule-guided evidence selection.}
    DelSA explicitly reorganizes temporally disordered information and selects evidence through self-updating rules, allowing the model to better identify temporal conflicts and reduce the influence of irrelevant evidence. The consistent improvements across all metrics show that DelSA provides a more reliable solution for temporal fake news detection.
\end{itemize}
\subsection{Ablation Study}

\begin{figure}[htbp]
  \centering
  \includegraphics[width=\linewidth]{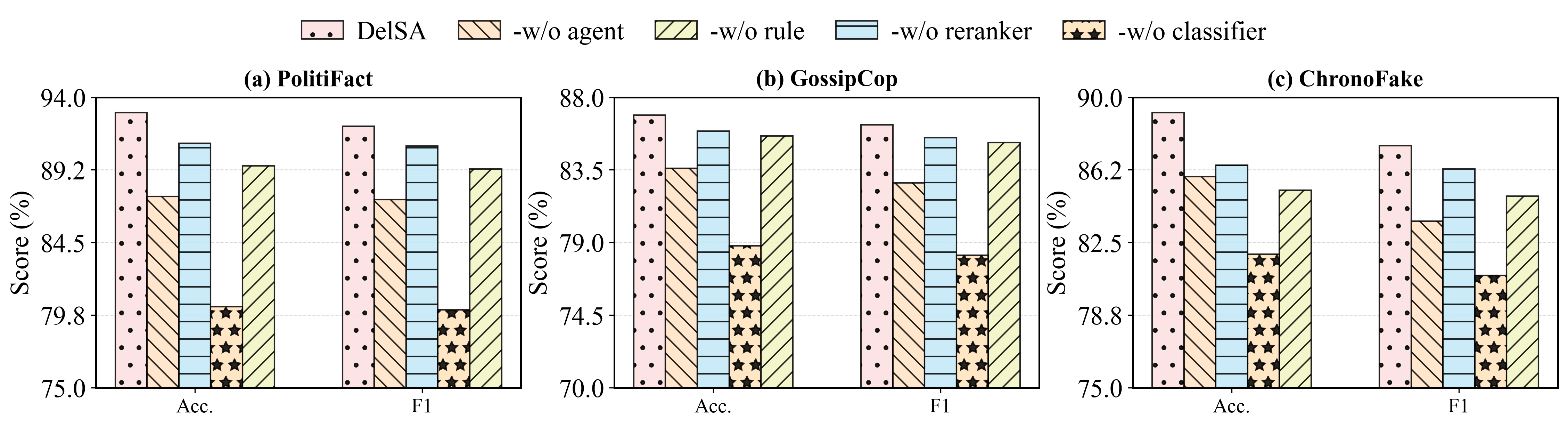}
  \caption{Ablation study results of the proposed fake news detection model on three datasets.}
  \label{fig:ablation}
\end{figure}

To assess the contribution of different components in \textbf{DelSA}, we compare the full model with four ablated variants. \textit{w/o agent} removes the time-aware agent and directly uses the initialized event--evidence pairs for classification. \textit{w/o rule} removes the rulebook guidance from evidence selection. \textit{w/o reranker} disables the MLP-based reranker and lets the LLM select evidence directly. \textit{w/o classifier} replaces the final supervised classifier with direct LLM-based verification.

For reference, the full DelSA model achieves weighted F1 scores of 0.924, 0.863, and 0.893 on PolitiFact, GossipCop, and ChronoFake, respectively, as reported in Tables~\ref{tab:performance} and~\ref{tab:disorder}. Figure~\ref{fig:ablation} summarizes the component-wise ablation trends. All ablated variants underperform the full model, indicating that each component contributes to the final performance. The main observations are summarized as follows:

\begin{itemize}
    \item \textbf{Effect of the time-aware agent.} Removing the agent causes a clear performance drop, indicating that adaptive temporal repair is important for resolving event--evidence misalignment.

    \item \textbf{Effect of rulebook guidance.} The degradation of \textit{w/o rule} shows that reusable selection rules help the LLM handle multi-time-point evidence ambiguity more reliably.

    \item \textbf{Effect of the reranker.} Without the MLP-based reranker, evidence selection becomes less stable because the model loses learned calibration over the LLM-generated quality scores.

    \item \textbf{Effect of the final classifier.} Replacing the supervised classifier with direct LLM verification weakens the integration of event-level evidence and article-level temporal structure, leading to less reliable predictions.
\end{itemize}
\subsection{Sensitivity Analysis}
\begin{figure}[htbp]      
  \centering              
  \includegraphics[width=\linewidth]{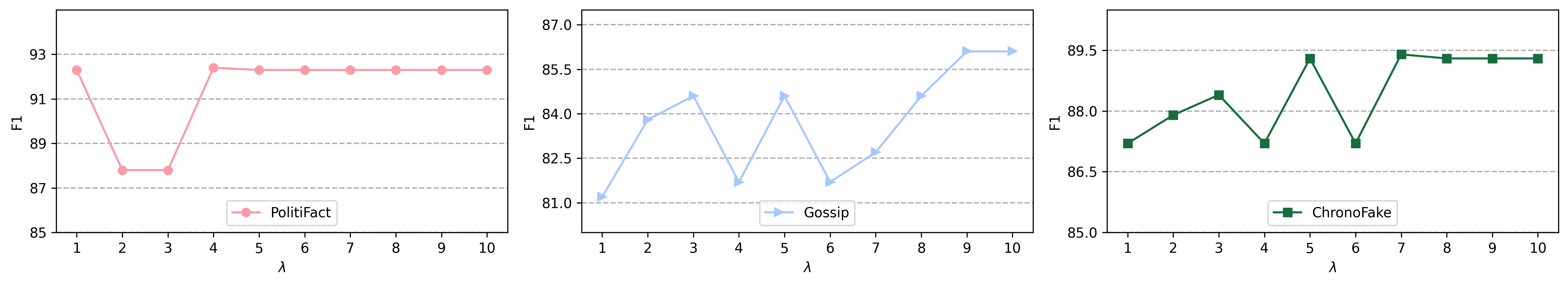}
  \caption{Sensitivity analysis of DelSA with respect to the number of update rounds on PolitiFact, GossipCop, and ChronoFake. The x-axis denotes the maximum number of training-time evidence update rounds, and the y-axis reports F1-score.}   
  \label{fig:param}        
\end{figure}

To investigate the influence of iterative evidence update, we vary the maximum number of update rounds from 1 to 10 while keeping all other hyperparameters fixed. For each setting, we train and evaluate DelSA on PolitiFact, GossipCop, and ChronoFake, and report the resulting F1 scores in Figure~\ref{fig:param}.

Across the three datasets, DelSA generally maintains stable performance under different update-round settings, indicating that the proposed evidence update mechanism is not highly sensitive to this hyperparameter. The detailed observations are as follows:

\begin{itemize}
    \item On PolitiFact, the model achieves stable performance after a small number of update rounds. Although slight performance drops appear when the update depth is too shallow, the F1-score quickly recovers and remains nearly unchanged as the number of rounds increases. This suggests that PolitiFact requires only limited evidence correction to obtain reliable event-evidence alignment.

    \item On GossipCop, the performance fluctuates more noticeably across different update rounds. This may be because GossipCop contains more diverse topics and a noisier evidence space, making the evidence reselection process more sensitive to the update depth. Nevertheless, the overall performance improves when more update rounds are allowed, showing that additional evidence adjustment can help the model handle complex evidence ambiguity.

    \item On ChronoFake, DelSA consistently achieves strong performance across most settings, especially when the number of update rounds is moderately large. This demonstrates that iterative evidence update is particularly useful for temporally disordered fake news detection, where the model needs to repeatedly adjust event-evidence assignments to reconstruct a more reliable evidence-grounded timeline.
\end{itemize}

\subsection{Out-of-Distribution Testing}
\begin{table}[t]
\centering
\caption{Performance comparison in the out-of-distribution setting. We employ a leave-one-dataset-out protocol in which the model is trained on two domains and tested on the third unseen domain shown in the first column. Best results are bolded, and second-best results are underlined.}
\label{tab:ood_results}
\begin{tabular}{llcccc}
\toprule
\textbf{Domain} & \textbf{Method} & \textbf{Acc.} & \textbf{Pre.} & \textbf{Rec.} & \textbf{F1} \\
\midrule
\multirow{4}{*}{PolitiFact}
& DRES     & \textbf{0.821} & \textbf{0.827} & \textbf{0.821} & \textbf{0.797} \\
& SheepDog & 0.638 & 0.727 & 0.642 & 0.593 \\
& MUSER    & 0.535 & 0.678 & 0.535 & 0.577 \\
& DelSA    & \underline{0.789} & \underline{0.756} & \underline{0.789} & \underline{0.772} \\
\midrule
\multirow{4}{*}{GossipCop}
& DRES  & \underline{0.661} & \textbf{0.775} & \underline{0.661} & \underline{0.665} \\
& SheepDog & 0.520 & 0.604 & 0.520 & 0.522 \\
& MUSER    & 0.517 & 0.582 & 0.517 & 0.527 \\
& DelSA & \textbf{0.705} & \underline{0.711} & \textbf{0.705} & \textbf{0.700} \\
\midrule
\multirow{4}{*}{ChronoFake}
& DRES  & \underline{0.664} & \textbf{0.557} & \underline{0.664} & \underline{0.567} \\
& SheepDog & 0.643 & 0.520 & 0.643 & 0.536 \\
& MUSER    & 0.554 & \underline{0.542} & 0.554 & 0.548 \\
& DelSA & \textbf{0.697} & 0.535 & \textbf{0.697} & \textbf{0.589} \\
\bottomrule
\end{tabular}
\end{table}
To assess the generalizability of \textbf{DelSA} across different fake news domains, we further extend the evaluation to a leave-one-dataset-out setting on the three datasets, including \textit{PolitiFact}, \textit{GossipCop}, and \textit{ChronoFake}. In each run, we preserve one dataset as the unseen test set, while the samples from the other two datasets are combined for training and validation. This setting allows us to evaluate whether DelSA can generalize to an unseen domain with different data distributions and temporal characteristics.
The results are summarized in Table~\ref{tab:ood_results},
and key findings are as follows:
\begin{itemize}
    \item DelSA achieves the best out-of-distribution performance on GossipCop and ChronoFake, showing strong generalization to unseen domains with different data distributions and temporal characteristics.

    \item On ChronoFake, DelSA obtains the highest accuracy and F1-score, indicating that its temporal evidence selection mechanism remains effective even when temporally disordered samples are not observed during training.

    \item On PolitiFact, DelSA is slightly behind DRES but outperforms the LLM-based and retrieval-augmented baselines, suggesting that it remains competitive on conventional fake news benchmarks under domain shift.

    \item MUSER shows limited generalization across all three unseen domains, implying that retrieval-augmented verification can be brittle when evidence selection is not explicitly guided by temporal compatibility.

    \item These results demonstrate that DelSA improves cross-domain robustness by selecting temporally reliable evidence rather than relying only on surface semantics or local evidence support.
\end{itemize}

\subsection{Case Study}
\begin{figure}[t]      
  \centering              
  \includegraphics[width=\linewidth]{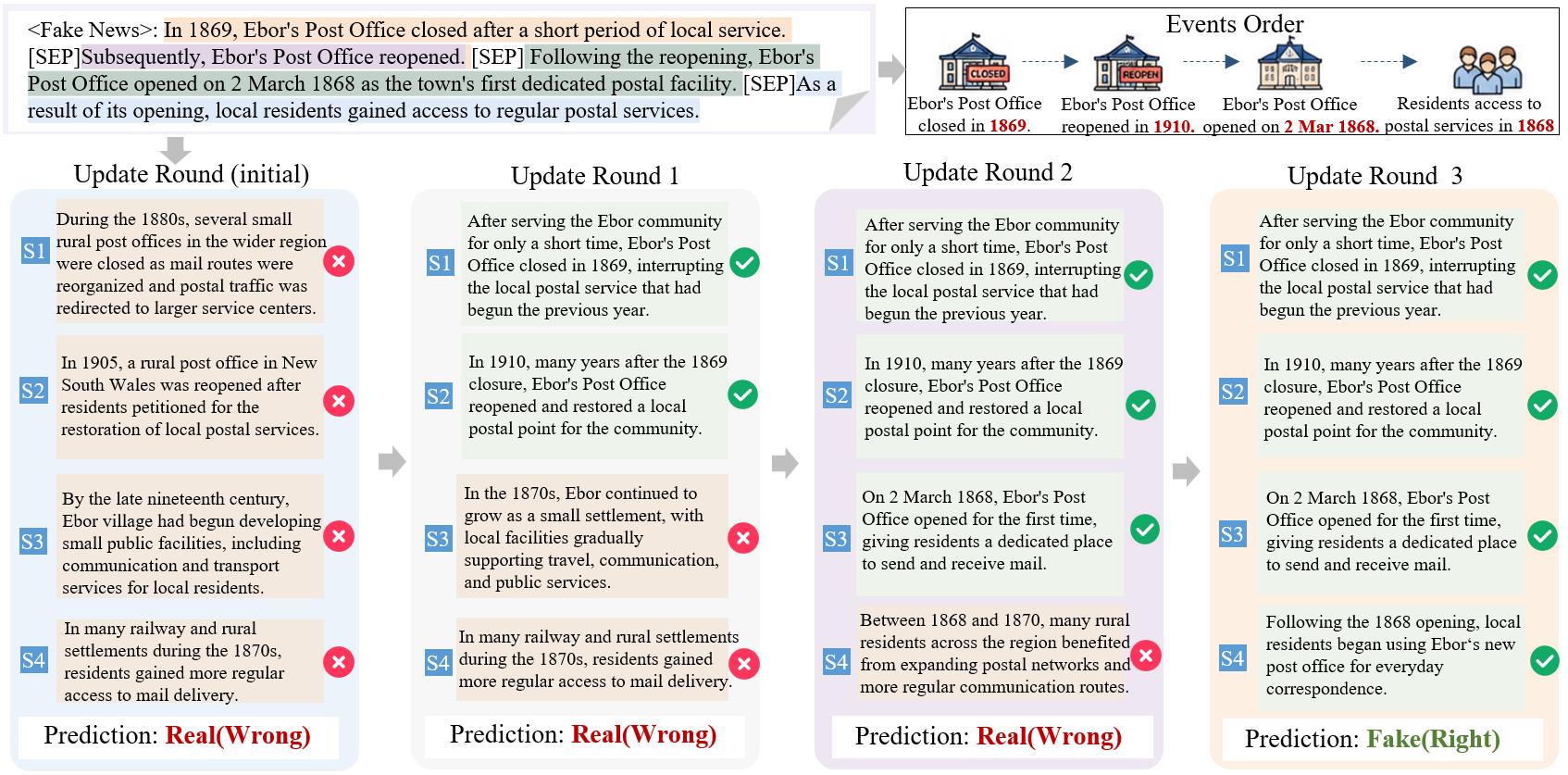}
  \caption{A case study of DelSA on temporally disordered fake news.}   
  \label{fig:case}        
\end{figure}
Figure~\ref{fig:case} presents a successful case in which DelSA corrects an initially wrong prediction through training-time iterative evidence selection. The input article is centered on Ebor's Post Office, but it constructs an incorrect temporal narrative by placing the 1868 opening after the 1869 closure and the 1910 reopening. At the initial training state, the selected evidence is only superficially relevant: it discusses regional post-office closures, a rural post-office reopening, and general postal access, but does not directly verify the target events or their timestamps. Because these evidence pieces are semantically coherent yet temporally insufficient, the model fails to expose the chronological conflict and predicts the article as real.

Across subsequent training updates, DelSA progressively repairs the event--evidence alignment. It first selects evidence directly supporting the 1869 closure and 1910 reopening of Ebor's Post Office, then identifies the explicit opening date of 2 March 1868, and finally grounds the service-access event in the 1868 opening. After this update process, the reconstructed evidence timeline indicates that the office opened in 1868, enabled residents to access postal services, closed in 1869, and reopened in 1910. In contrast, the article claims that the closure and reopening occurred before the initial opening and service access. This mismatch reveals the temporal manipulation and changes the prediction from real to fake, showing how DelSA escapes the evidence coherence trap by replacing locally relevant but temporally inappropriate evidence with time-consistent evidence.
\section{Conclusion}
This paper addresses the challenge of temporally disordered fake news, where real events are presented out of chronological order to mislead readers.
It introduces DelSA, an LLM-driven, rule-guided framework that refines event-level evidence during training, reconstructs an evidence-grounded timeline, and applies a frozen rulebook for one-pass evidence selection at inference.
We also construct ChronoFake, a large-scale benchmark designed to evaluate models on temporally manipulated narratives.
Experimental results show that DelSA outperforms the evaluated content-based, LLM-enhanced, and retrieval-augmented baselines on ChronoFake while remaining competitive on PolitiFact and GossipCop, highlighting the importance of temporal reasoning and evidence-grounded verification in detecting subtle fake news.


\bibliographystyle{ACM-Reference-Format}
\bibliography{sample-base}

\appendix

\end{document}